\documentclass[10pt,a4paper]{article}
\usepackage{lrec2022} 
\usepackage{multibib}
\usepackage{graphicx}
\usepackage{tabularx}
\usepackage{soul}
\usepackage[dvipsnames]{xcolor}
\usepackage{todonotes}
\usepackage{comment}
\usepackage{epstopdf}
\usepackage[utf8]{inputenc}
\usepackage{hyperref}
\usepackage{xstring}
\usepackage[autostyle, english = american]{csquotes}
\MakeOuterQuote{"}
\usepackage{titlesec}
\titleformat{\section}{\normalfont\large\bfseries\center}{\thesection.}{1em}{}
\titleformat{\subsection}{\normalfont\SmallTitleFont\bfseries\raggedright}{\thesubsection.}{1em}{}
\titleformat{\subsubsection}{\normalfont\normalsize\bfseries\raggedright}{\thesubsubsection.}{1em}{}
\renewcommand\thesection{\arabic{section}}
\renewcommand\thesubsection{\thesection.\arabic{subsection}}
\renewcommand\thesubsubsection{\thesubsection.\arabic{subsubsection}}

\usepackage{color}
\usepackage{wasysym}

\title{Features of Perceived Metaphoricity on the Discourse Level:\\Abstractness and Emotionality\\ \vspace*{.5\baselineskip}} 

\name{Prisca Piccirilli, Sabine Schulte im Walde} 

\address{Institute for Natural Language Processing (IMS), University of Stuttgart\\
         Pfaffenwaldring 5b, 70569 Stuttgart, Germany\\
         \{prisca.piccirilli,schulte\}@ims.uni-stuttgart.de\\}

\abstract{ 
Research on metaphorical language has shown ties between abstractness and emotionality with regard to metaphoricity; prior work is however limited to the word and sentence levels, and up to date there is no empirical study establishing the extent to which this is also true on the discourse level. This paper explores which textual and perceptual features human annotators perceive as important for the metaphoricity of discourses and expressions, and addresses two research questions more specifically. First, is a metaphorically-perceived discourse more abstract and more emotional in comparison to a literally-perceived discourse? Second, is a metaphorical expression preceded by a more metaphorical/abstract/emotional context than a synonymous literal alternative?
We used a dataset of 1,000 corpus-extracted discourses for which crowdsourced annotators (1)~provided judgements on whether they perceived the discourses as more metaphorical or more literal, and (2)~systematically listed lexical terms which triggered their decisions in (1). Our results indicate that metaphorical discourses are more emotional and to a certain extent more abstract than literal discourses. However, neither the metaphoricity nor the abstractness and emotionality of the preceding discourse seem to play a role in triggering the choice between synonymous metaphorical vs. literal expressions. Our dataset is available at \url{https://www.ims.uni-stuttgart.de/data/discourse-met-lit}.
 \\ \newline  }

\begin{document}

\maketitleabstract

\section{Introduction}
\label{intro}

Metaphors represent a "necessary, not just nice" element of everyday thought and communication  \cite{Ortony:1975,Lakoff-Johnson:1980,vandenBroek:1981,Schaeffner:2004}, as the underlying motivation of metaphorical language usage is to facilitate the understanding of a more abstract target domain through linguistic expressions of a more concrete source domain \cite{Lakoff-Johnson:1980}. 
For example, the metaphorical sentences \textit{I am \underline{at a crossroads} in my life} or \textit{He's never let anyone \underline{get in his way}}, illustrate how the abstract concept of \textit{life} in English is expressed through the more concrete concept of a \textit{journey}. 

More than a simple stylistic way of expressing one's thoughts, metaphors are a powerful instrument to facilitate communication, as they are used to explain and/or explicate things \cite{Kirklin:2007,Stefanowitsch:2008} that are unfamiliar or unknown \cite{Glucksberg:1989}, but also in a more efficient way \cite{Ortony:1975}. They have proven to shape thinking and influence reasoning \cite{Boroditsky:2011,Thibodeau-Boroditsky:2011,Thibodeau-etal:2017}, and they may also represent an effective persuasive device \cite{VanStee:2018}.  

According to corpus-linguistic studies, metaphors manifest themselves in general-domain text corpora in every third sentence
\cite{Gedigian-etal:2006,Shutova-Teufel:2010,Steen-etal:2010}, which makes the processing of metaphorical language primordial for many natural language processing ({\small NLP}) tasks and applications.
The vast majority of work in {\small NLP} has been concerned with the \textit{detection} of metaphorical expressions\footnote{\newcite{Tong-etal:2021} offer a systematic, comprehensive review and discussion of the most recent metaphor processing systems and datasets.}. 
Linguistically and conceptually-driven strategies involve identifying selectional preference violations (compare \textit{consume food} vs. \textit{consume information}) \cite{Fazly-etal:2009,Shutova-etal:2013,Ehren-etal:2020}, judging discourse coherence \cite{Sporleder-Li:2009,Bogdanova:2010,Dankers-etal:2020}, and inducing
discourse features indicating figurative language, such as supersenses, concreteness, emotionalty, imageability \cite{Turney-etal:2011,Tsvetkov-etal:2013,Koeper-SchulteImWalde:2016b,Mohammad-etal:2016,Koeper-Schulte-im-walde:2018,Alnafesah-etal:2020,Hall-etal:2020}. 

\begin{table*}[ht]
\begin{tabular}{p{0.97\linewidth}}
\hline
\textbf{This wasn't just a play on words, rather it was a demand that they should 'maintain a consistency between their words and their actions'. But I agree, that still does not absolve them from the need to speak truth to power. In our times when people spend so much time with TV and the internet, do they have the interest and time to read poetry? Many people believe that it is difficult to read poetry. Can everyone} [\textcolor{red}{grasp} / \textit{\color{YellowOrange}{understand} the meaning} of a good poem, or is a skill necessary?]\\
\hline
\end{tabular}
\caption{Example of a \textbf{discourse} collected by \protect\newcite{Piccirilli-SchulteImWalde:2021}. 
The part in brackets was not presented to the annotators, to obtain judgements solely on the discourse preceding the use of the target expression.}
\label{tab:example-pair}
\vspace{-0.9mm}
\end{table*}

Existing research has however focused solely on sentences with word-level and phrase-level metaphorical expressions, and there has been little discussion on metaphoricity on the discourse level. 
Although the cognitive and psycholinguistics communities working on metaphorical language have provided strong evidence for the importance of context \cite{Inhoff-etal:1984,Giora:2003,Martin:2008,Koevecses:2009}, only very few recent {\small NLP} studies have incorporated wider discourse properties into their models \cite{Mu-etal:2019,Dankers-etal:2020,Gong-etal:2020}.
As a result, we still have a very limited understanding of \textbf{what actually constitutes a metaphorically-perceived discourse}. Metaphor processing systems would benefit from both a better understanding of the cognition of metaphor and its role and effect in communication. 

In the current study, we overcome this limited perspective, making use of a  dataset of 1,000 corpus-extracted discourses that we collected in previous work (see an example in Table \ref{tab:example-pair}), where we asked participants to rate the degree of discourse literalness vs. metaphoricity on a scale from 1 (mostly literal) to 6 (mostly metaphorical) \cite{Piccirilli-SchulteImWalde:2021}\footnote{Henceforth we refer to this previous work as P\&SiW21.}. 
From a theoretical point of view, our rating scale approach was highly appropriate, given that a discourse as a whole is rarely "purely metaphorical" or "purely literal". We were however concerned from an annotation point of view, given that judgements on discourse metaphoricity represent a difficult, challenging and subjective task, where various discourse properties may interact and influence the annotators' decisions. We thus provide a more specific task in the current annotation setup, to answer the question of \textbf{what is driving annotators to perceive a discourse more metaphorical or more literal}.

In this paper, we rely on our previous collected discourses in P\&SiW21, and zoom  into lexical features that potentially influence the perception of a given discourse as more or less metaphorical. 
First, contrarily to using rating scales for annotating the metaphoricity of the discourses, we ask annotators to proceed to a binary decision, whether they perceive the discourses more literal or more metaphorical. This step enables us to compare our findings to those of P\&SiW21 in order to (i) explore whether asking the same question in different tasks varies the output (ratings vs. binary decision), and (ii) confirm or not the context-salient hypothesis \cite{Koevecses:2009}, i.e., that a metaphorical vs. literal discourse precedes a metaphorical vs. literal expression, respectively.
Then, we ask annotators to provide a list of five unique words which triggered their decision. 
By systematically collecting the lexical items after a binary judgement on the discourse, we hope for a more reliable annotation collection, as annotators are forced to reflect and re-read the discourse to fully complete the tasks. Moreover, by relying on existing English lexicons for abstractness and emotionality ratings, we also shed light on the lexical conditions of abstractness and emotionality for metaphorically-perceived discourses, by relating our findings to the judgements on overall discourse metaphoricity. Finally, these ratings enable us to explore abstractness and emotionality as features for the choice between synonymous metaphorical vs. literal expressions. 

\section{Related Work}
\label{relwork}

\paragraph{Annotations of Metaphoricity}

The {\small NLP} tasks of figurative language \textit{identification} and \textit{interpretation} have led to the creation of several datasets on the metaphoricity of \textit{lexical items}.
The most largely used resource is the {\small VU} Amsterdam Metaphor Corpus \cite{Steen-etal:2010}, comprising judgements on 187,570 terms from the {\small BNC} \cite{BNC:2007}.

Other datasets aim to explore the choice between a metaphorical expression vs. a synonymous literal counterpart. Most prominently, \newcite{Mohammad-etal:2016} composed 171 sentences where a verb is used metaphorically, e.g., \textit{abuse} in "Her husband often abuses alcohol". For each sentence, the authors of the paper chose a literal synonym of the target verb, such as \textit{drink} in the above example.
\newcite{Shutova:2010} collected sentences containing metaphorically-used verbs from the {\small BNC}, e.g., \textit{grasp} in "Anyone  who  has introduced  speech  act  theory  to  students  will  know  that  these  technical  terms [...] are  not  at  all easy  to  grasp." Annotators were asked to provide an alternative literally-used verb. The dataset consists of a list of metaphorically-used verb--object ({\small VO}) and subject--verb ({\small SV}) expressions, with one or more literal verb alternatives.
For instance, the verb \textit{grasp} in the {\small VO} expression \textit{grasp term} was related to the literal alternatives \textit{understand} and \textit{comprehend}.
\newcite{Bizzoni-Lappin:2018} developed a system which automatically ranks the best four paraphrases for metaphorical sentences. Their dataset contains 200 metaphorical sentences, each with their automatically generated and ranked paraphrases\footnote{\url{https://github.com/yuri-bizzoni/Metaphor-Paraphrase/}}.

These impressive resources however show two common limitations: (i)~the annotation of metaphoricity is \textit{word-based}, and (ii)~the target items are in only \textit{one-sentence contexts}, which is not sufficient when addressing metaphoricity on the \textit{discourse level}.   
We filled this gap in P\&SiW21. Based on 50 pairs of English synonymous literal vs. metaphorical {\small VO} and {\small SV} expressions from \newcite{Shutova:2010} and \newcite{Mohammad-etal:2016}, we introduced the above-mentioned dataset with 1,000 discourses of four to five sentences, where both expressions from a synonymous pair can be used in the final sentence, as shown in the example in Table \ref{tab:example-pair}. Annotators were asked, in a two-step crowdsourcing experiment, to (1) rate the degree of metaphoricity of the discourse,  and (2) choose the expression that fits best. 

\paragraph{Metaphor and Abstractness}

In their metaphor identification procedure ({\small MIP}), \newcite{Steen-etal:2010} define a word as metaphorical or literal based on whether it has a "more basic meaning" in other contexts than the current one; "basic meaning" here is defined as "more concrete". 
After all, research in cognitive linguistics views metaphor as a figurative device for transferring knowledge from a concrete domain to a more abstract domain  \cite{Lakoff-Johnson:1980}, as exemplified by the conceptual metaphor {\small LIFE IS A JOURNEY} in the introduction.

This view led to the hypothesis that metaphorical word usage is correlated with the degree of abstractness of the word's contexts, and this hypothesis has been supported in numerous studies where abstractness features have proven useful for automated metaphor \textit{identification} \cite{Turney-etal:2011,Tsvetkov-etal:2013,Koeper-SchulteImWalde:2016b,Alnafesah-etal:2020,Hall-etal:2020}.

\paragraph{Metaphor and Emotion}

The concept of affect has also been explored with regard to metaphoricity. 
A number of studies not only have found that metaphorical language has a stronger emotional effect than literal language in general \cite{Blanchette-Dunbar:2001,Crawford:2009}, but also that metaphorical words and sentences are more emotionally engaging than their synonymous literal counterparts \cite{Citron-Goldberg:2014,Mohammad-etal:2016}.

In {\small NLP}, informing models with emotionality features has also proven to be useful for metaphor detection. While \newcite{Garget-Barnden:2015} and \newcite{Koeper-Schulte-im-walde:2018} showed that emotion features are improving performance in distinguishing metaphorical vs. literal usages, \newcite{Dankers-etal:2019} found that dominance (the perceived degree of control in a social situation) also improved considerably their {\small BERT} model to identify the use of metaphors.

\paragraph{Role of Context for Metaphorical Language}
On the one hand, many words are ambiguous, and the empirical study provided by \newcite{Mohammad-etal:2016} confirmed the hypothesis that variation in metaphorical and literal language usage is a common pattern for polysemous verbs. It is thus very probable that a metaphorically-used lexical item has a literal counterpart, and vice versa. 
On the other hand, the cognitive and psycho-linguistics communities have shown the powerful instrument that metaphorical language is \cite{Boroditsky:2011,Thibodeau-etal:2017,VanStee:2018}.
Both of these aspects therefore call for the necessity to go beyond word- and sentence-based approaches and look at larger contexts when one explores metaphorical language. 

\newcite{Koevecses:2009} discusses contextual factors that play a role in the production of metaphors in natural discourse, such as the linguistic context itself, the major entities participating in the discourse, the physical and social settings as well as the cultural context. 
Recently, some work informed their models with discourse properties. \newcite{Mu-etal:2019} used general-purpose word, sentence and document embedding methods, and \newcite{Dankers-etal:2020} used hierarchical attention to their fine-tuned {\small BERT} model, and both studies improved metaphor identification, illustrating the importance of context beyond sentence. 

\section{Datasets and Lexicons}
\label{data}

\paragraph{Metaphoricity Ratings} Based on 50 pairs of English synonymous literal vs.~metaphorical {\small VO} and {\small SV} expressions from previous work such as \textit{understand} vs. \textit{grasp meaning} \cite{Shutova:2010,Mohammad-etal:2016}, we collected and analyzed human judgements on 1,000 corpus-extracted discourses of four to five sentences, where both the metaphorical and literal target expressions can be used in the last sentence, as in Table \ref{tab:example-pair}, cf. P\&SiW21. Using the crowdsourcing platform Amazon Mechanical Turk (MTurk)\footnote{\url{https://www.mturk.com/}}, each discourse was annotated by at least 11 workers regarding the degree of metaphoricity of the discourse, on a scale from 1 (mostly literal) to 6 (mostly literal). To the best of our knowledge, this dataset constituted the first solid starting point to further explore salient discourse conditions for contextual metaphorical vs. literal usage, and we therefore use the collected discourses for the present work. 

\paragraph{Abstractness Lexicon} The lexicon from \newcite{Brysbaert-etal:2014} provides abstractness ratings for 37,058 English words. The data was collected on the basis of the {\small SUBTLEX-US} corpus \cite{Brysbaert-New:2009}, the English Lexicon Project \cite{Keuleers-etal:2011}, the corpus of contemporary American English \cite{Davies:2009}, and additional words that they personally gathered. They recruited participants via Mturk, who were residents of the United States and English native speakers. Each word was annotated by at least 25 participants, on a scale from 1 (= more abstract) to 5 (= more concrete). 
We use these ratings to measure the overall abstractness score for each of our annotated discourses.

\paragraph{Emotionality Lexicon} \newcite{Buechel-etal:2020} introduced a new methodology to automatically generate lexicons for 91 languages comprising eight emotional variables: \textit{Valence, Arousal, Dominance} ({\small VAD}) as well as the five basic emotions \textit{Joy, Anger, Surprise, Fear, Surprise} ({\small BE5}) \cite{Ekman:1992}. As a source dataset, they used the English emotion lexicon from \newcite{Warriner-etal:2013}, comprising about 14K entries in {\small VAD} format collected via crowdsourcing. They applied the {\small BE5} ratings from \newcite{Buechel-Hahn:2018a} to convert the {\small VAD} ratings. Via their monolingual state-of-the-art multi-task feed-forward network \cite{Buechel-Hahn:2018b}, they projected ratings on these eight variables and resulted with an English lexicon of 2M word type entries with very high correlation with human judgements (around 90\% for each variable). 
We use the {\small BE5} ratings of the English lexicon for our study.

\section{Experiment Setup and Hypotheses}

\subsection{Crowdsourcing Experiment}
\label{subsec:experiment}
\paragraph{Annotation Scheme} Contrarily to using rating scales for annotating the metaphoricity of the discourses and then obtaining a metaphorical vs. literal categorization for each discourse using 3.5 as a threshold  (P\&SiW21), we provided a more specific task in the current annotation setup and asked annotators to proceed to a binary decision, whether they perceived the discourses as \textit{more literal} or \textit{more metaphorical}. 
Note that similarly to P\&SiW21, we deliberately did not give any explanations to the annotators with regard to what a metaphorical vs. literal discourse is. Previous cognitive linguistic research has shown how metaphors are used everyday at a large scale by everybody, and in an unintentional way \cite{Lakoff-Johnson:1980}: we did not wish to bias the annotators and we therefore let their responses be driven solely by their intuition on what constitutes a metaphorical vs. literal discourse. 

For each discourse, ten annotators were asked to read the discourse and judge whether it was \textit{more literal} or \textit{more metaphorical} (Task~1). 
In order to avoid a bias of the annotators since the original discourses contain a target metaphorical or literal expression in the last sentence (see Table \ref{tab:example-pair}), we presented the discourses up to the word preceding the expression, as in P\&SiW21.
This approach enables us to compare the findings.

As the present study is driven by the interest to zoom into discourse features that potentially influence the perception of a given discourse as more or less metaphorical, annotators were asked to reflect on their choice.
After providing a binary decision with regard to the metaphoricity of the discourse, they were asked to \textit{provide five unique words} that triggered their perception of the discourse being more metaphorical or literal (Task~2). 
We provide in Appendix \ref{app:crowd} an example of what a MTurk worker would annotate when providing judgement for one discourse.

We limited the location of the workers to English-speaking countries, and specified in the instructions that the tasks were only for English native speakers.
We paid each annotator \$0.03 per judgement. We refused judgements if (i)~Task~2 was not completed at all, (ii)~they had provided less than three words for a judgement in Task~2 and (iii)~they had provided complete chunks of texts instead of unique words in Task~2. 

\paragraph{Decisions and Study Set} Each of our initial 1,000 discourses was annotated by ten MTurk workers, resulting in a dataset of 10,000 judgements, from a total of 688 workers who provided on average 14 judgements. 
However, for our analyses, we choose a strict criterion to obtain greater confidence on our results. 
First, we only keep the judgements of workers who annotated at least 20 discourses in total. Then, for an instance to be considered metaphorical or literal, 70\% or more of the annotators of that instance had to agree on the choice of the category. The instances for which this level of agreement was not reached are discarded from further analyses. 
The subset of the dataset we analyze for this study therefore consists of 5,844 instances, annotated by 200 MTurk workers who provided on average 41 judgements. 
For each of the 5,844 instances, annotators listed five lexical items which triggered their choice of the category, but we also keep 32 instances for which three and four words were provided.

When combining all the annotated information for each of our 1,000 discourses and after proceeding to this strict decisions we describe above, our \textit{study set} results in 711 discourses, for which a minimum of 70\% of the annotators agreed on the choice of category (metaphorical or literal discourse), and with an average of 41 tokens to support the choice of category.
Even though we discarded instances for this study to reach better reliability, we release the full set of 10,000 annotated instances (1,000 discourses) for other uses and further research, next to the study set.

\subsection{Abstractness Ratings}
\label{subsec:abstr-ratings}

Across all discourses, the annotators provided an average of 41 tokens to indicate what triggered their metaphorical or literal perception of the respective discourse.
Based on the abstractness lexicon from \newcite{Brysbaert-etal:2014} (Section \ref{data}), each individual token is assigned its abstractness rating. Note that we consider all tokens (and not types), as we believe that the "abstractness load" of a lexical item which was picked more than once by several annotators of the same discourse should weigh more.
We then compute the average score for each discourse to obtain an overall rating of abstractness. 

Previous work has shown that part-of-speech tags play an important role in assigning abstractness ratings.
Indeed, nouns are considered to be easier to evaluate by humans and their abstractness scores are thus considered more reliable \cite{Koeper-Schulte-im-walde:2016,Frassinelli-Schulte-im-walde:2019,SchulteImWalde-Frassinelli:2022}; they also tend be perceived as more concrete and therefore obtain higher concreteness scores \cite{Brysbaert-etal:2014,Frassinelli-Schulte-im-walde:2019,SchulteImWalde-Frassinelli:2022}; and word classes differ regarding their strength of perception across senses \cite{Lynott/Connell:13,Connell/Lynott:15}.
We therefore consider four distinct settings to assign an abstractness score to each discourse: where \textbf{all} picked tokens ({\small ALL}), only \textbf{nouns} ({\small N}), only \textbf{verbs} ({\small V}), and only \textbf{adjectives} ({\small A}) are mapped to their individual abstractness ratings.
If the context's degree of abstractness indeed applies to the overall metaphoricity of that said context, we expect to observe that: 

\textbf{Hypothesis 1a:} Metaphorically-perceived discourses are more abstract/less concrete.
We expect this observation even more so in settings ({\small V}) and ({\small A}), as verbs and adjectives are the most frequent type of metaphor \cite{Shutova-Teufel:2010,Gandy-etal:2013}, and tend to be perceived more abstract than nouns \cite{Koeper-Schulte-im-walde:2016}.

\textbf{Hypothesis 1b:} With regard to the choice between synonymous metaphorical vs. literal expressions, we expect discourses in which a metaphorical expression is used to be preceded by more abstract context, but in which a literal expression is used to be preceded by a more concrete context.

\subsection{Emotionality Ratings}
\label{subsec:emo-ratings}

Similarly to Section \ref{subsec:abstr-ratings}, we assign each discourse an overall emotionality score, based on the English emotion lexicon from \newcite{Buechel-etal:2020} (Section \ref{data}).
The emotion lexicon provides ratings for {\small VAD} as well as {\small BE5}. For the present study, we are interested in the emotionality of the picked lexical items, i.e., the emotional load that a term conveys, rather than the actual emotion a term refers to. 
Out of the five scores that a term receives -- one for each emotion, we assume that its highest score is reflective of the "emotional load" of that term, i.e., how much emotion it conveys. For example, assuming that the lexical item "truth" is picked for a certain discourse: it obtained the scores 2.24, 1.46, 1.40, 1.49, 1.46 for Joy, Anger, Sadness, Fear and Surprise, respectively, in the emotion lexicon. In our experiments, the term "truth" is therefore attributed the score of 2.24. 
Each lexical item for each discourse is attributed an emotionality score following this assumption, and each discourse then obtains one final score computed from the average of all items' scores.  
If the context's degree of emotionality indeed applies to the overall metaphoricity of that said context \cite{Crawford:2009,Citron-Goldberg:2014,KoustaEtAl:11}, we hypothesize that:\\
\textbf{Hypothesis 2a:} Metaphorically-perceived discourses are emotionally more loaded 
than literally-perceived discourses. 

\textbf{Hypothesis 2b:} With regard to expressions, we expect a more emotional discourse to be followed by the use of a metaphorical expression, whereas a less emotional discourse is expected to be followed by the use of a literal expression.

\section{Results and Data Analysis}
\label{sec:results}

We present the findings of our annotation studies, first addressing our research question regarding the perception of metaphoricity in \textbf{discourses} and the interaction with abstractness and emotionality (Section \ref{subsec:discourses}). We then address our second research question, focusing on the metaphoricity of \textbf{expressions} and the role of the preceding context in the lexical choice (Section \ref{subsec:expressions}).
\subsection{Discourse Properties}
\label{subsec:discourses}

\begin{figure}[t]
\begin{center}
\includegraphics[scale=0.25]{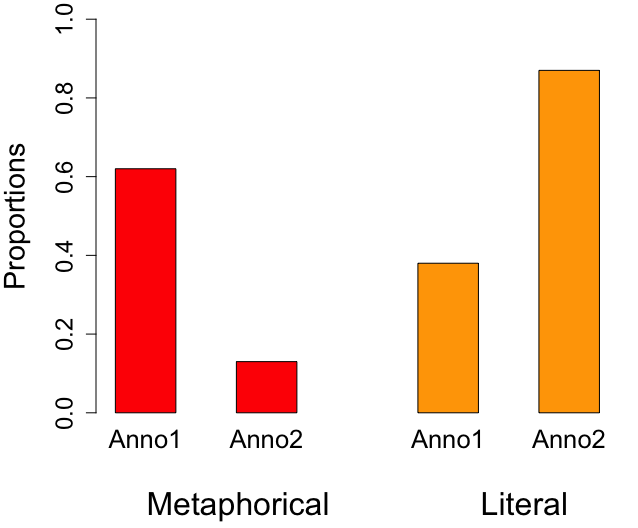} 
\caption{Proportions of metaphorically- vs. literally-perceived discourses (red vs. orange) in P\&SiW21 (Anno1) and the present study (Anno2).}
\label{fig:met-lit-proportions}
\end{center}
\end{figure}

\paragraph{Metaphoricity} 
P\&SiW21 collected human judgements on metaphorically vs. literally-perceived discourses by asking raters to judge on a scale 1--6 how metaphorical they perceived the discourses. They then used 3.5 as a threshold to categorize discourses as metaphorical or literal.
We asked the same question, but collected annotations as a binary categorization.
Figure \ref{fig:met-lit-proportions} provides a comparison of metaphorically- vs. literally-perceived discourses in P\&SiW21 (Anno1) and the present study (Anno2). 

\begin{figure*}[t]
\begin{center}
\includegraphics[height=50mm]{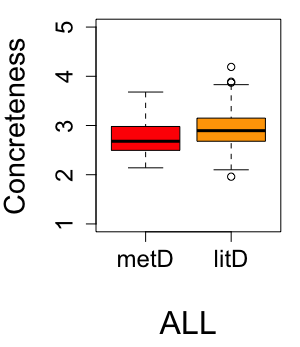} 
\includegraphics[height=50mm]{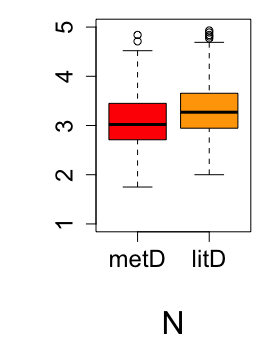} 
\includegraphics[height=50mm]{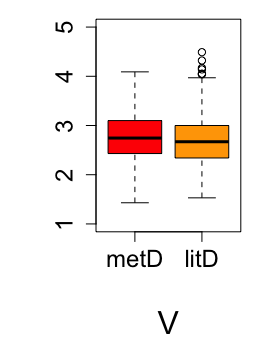} 
\includegraphics[height=50mm]{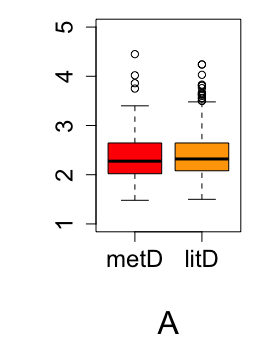} 
\caption{Abstractness of discourse w.r.t metaphorically (metD) vs. literally-perceived (litD) \textbf{discourses}, where 1 = more abstract, 5 = more concrete. Ratings are obtained from norms \protect\cite{Brysbaert-etal:2014} based on lexical items provided by annotators: all tokens (ALL), only nouns (N), only verbs (V) and only adjectives (A).}
\label{fig:discours-abstract}
\end{center}
\end{figure*}

While we previously found that humans judged more of the 1,000 discourses as metaphorical than literal (62.2\% vs. 37.8\% -- 622 vs. 378 discourses, respectively), we obtain opposite results in the present study, with our annotators perceiving 616 discourses as literal (86.6\%), and only 95 discourses as metaphorical (13.4\%).
Even though the question to collect human judgements is initially the same, i.e., how metaphorical a discourse is, these rather different proportions not only reveal the subjectivity of the task, but also the challenge to conduct the "right task" and consequently find the most reliable way to obtain human annotations.

However, it is worth mentioning those discourses where both studies agree on the metaphoricity or the literalness of discourses. As shown in Table \ref{tab:intersection}, almost half of the literally-perceived discourses in Anno2 were also literally-perceived in Anno1, and 85\% of the metaphorically-perceived in Anno2 were so in Anno1 as well, making those discourses highly reliable in terms of perceiving metaphoricity. We provide examples of these discourses in Appendix \ref{appsubsec:intersec}. 

\begin{table}[!h]
    \centering
    \begin{tabular}{l|r|r|r}
    \hline
        Perception & Anno1 & Anno2 & Anno1\&2  \\ \hline \hline
        Metaphorical & 622 & 95 & \textbf{81} \\
        Literal & 378 & 616 & \textbf{301} \\ \hline
    \end{tabular}
    \caption{Number of metaphorically- vs. literally-perceived discourses in P\&SiW21 (Anno1), in the present study (Anno2), and in the intersection of both collections (Anno1\&2).}
    \label{tab:intersection}
\end{table}

\paragraph{Abstractness} 
Figure \ref{fig:discours-abstract} offers an overview of the degree of abstractness of all discourses with respect to the binary metaphorical (in red) vs. literal (in orange) discourse judgements.

Hypothesis 1a assumes that metaphorically-perceived discourses are expected to be more abstract than literally-perceived discourses. We do observe this pattern in three of our four settings: ({\small ALL}), ({\small N}) and ({\small A}), however as a rather weak effect. 
For setting ({\small V}) the observation is reversed, even though this is precisely the setting for which we would have expected metaphorically-perceived discourses clearly more abstract than literally-perceived ones, given that verbs and adjectives are considered frequent types of metaphors \cite{Shutova-etal:2013,Gandy-etal:2013}.

Overall, our findings go against the tendency that we would have expected to observe, i.e., that metaphorically-perceived discourses \textit{clearly} tend to be more abstract than literally-perceived discourses. We do observe, however, that metaphorical discourses are more abstract with regard to nouns. 

\begin{figure}[h]
\begin{center}
\hspace*{-5mm} 
\includegraphics[scale=0.33]{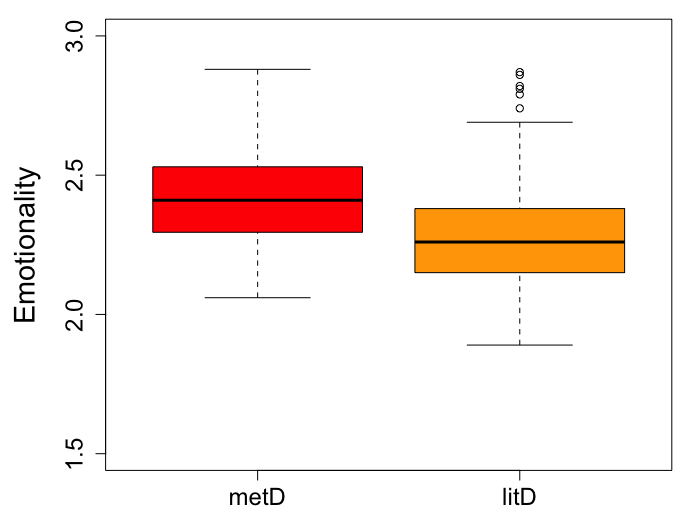} 
\caption{Emotionality of discourses w.r.t metaphorically (metD) vs. literally-perceived (litD)  \textbf{discourses}.}
\label{fig:discours-emotion}
\end{center}
\end{figure}

\paragraph{Emotionality} 
According to Hypothesis 2a, we expect metaphorically-perceived discourses to be more "emotionally-loaded" than literally-perceived discourses. 
We present in Figure \ref{fig:discours-emotion} the emotionality scores of our 711 discourses with respect to the binary metaphorical (in red) vs. literal (in orange) discourse judgements. 
Metaphorically-perceived discourses clearly seem to be emotionally more loaded, as they are assigned an average rating of 2.4, and the general scores do not drop below 2.0. The average emotionality rating for literally-perceived discourses lies at around 2.3, and for some instances, the score drops down to 1.7. Our results therefore support Hypothesis 2a, that metaphorically-perceived discourses also tend to be more emotional.

\subsection{Influence of Discourse Properties on Choice of Expressions}
\label{subsec:expressions}
We address in this section the extent to which metaphoricity, abstractness and emotionality play a role in the choice between synonymous metaphorical vs. literal \textbf{expressions}.

\paragraph{Metaphoricity} 
As exemplified in Table \ref{tab:example-pair}, the discourses of our dataset are composed of four to five sentences, where both a metaphorical expression or its synonymous literal counterpart are acceptable in the last sentence. 
Applying the context-salient hypothesis \cite{Koevecses:2009} to their study, P\&SiW21 tested whether a metaphorically-perceived discourse precedes a metaphorical expression, and ditto for a literally-perceived context and a literal expression. Their findings did not support this hypothesis. 

\begin{figure}[!h]
\vspace{+2mm}
\begin{center}
\hspace*{-4mm}  
\includegraphics[scale=0.38]{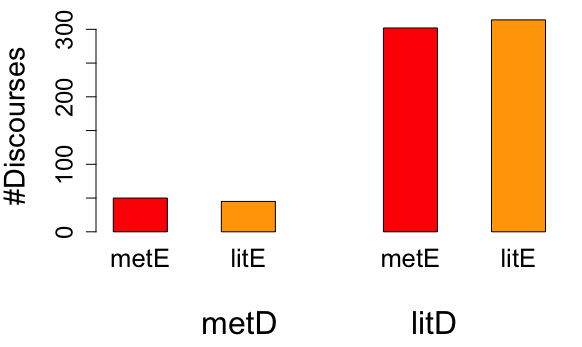} 
\vspace{+2mm}
\caption{Number of metaphorically (metD) and literally-perceived (litD) \textbf{discourses} followed by a metaphorical (metE) or literal (litE) \textbf{expressions}.}
\label{fig:discours-expressions}
\end{center}
\vspace{-3mm}
\end{figure}

As a matter of consistency, and since we obtained different results regarding the metaphoricity of discourses (\ref{subsec:discourses}), we also test the context-salient hypothesis. Figure \ref{fig:discours-expressions} provides the number of discourses that are perceived literal (litD) or metaphorical (metD) and followed by a literal expression (litE) vs. a metaphorical expression (metE). 

Our findings are aligned with those of P\&SiW21. 
We observe that metaphorical and literal expressions almost equally follow metaphorical and literal discourses, confirming that the metaphoricity of the preceding context does not seem to play a role in the usage of the following expression.

Similarly to what we did for the perception of discourses, we also look at the abstractness and emotionality of the preceding discourse as potential triggers for the choice between a synonymous metaphorical vs. literal expression and present our findings in the following two subsections.

\begin{figure*}[t]
\begin{center}
\includegraphics[height=50mm]{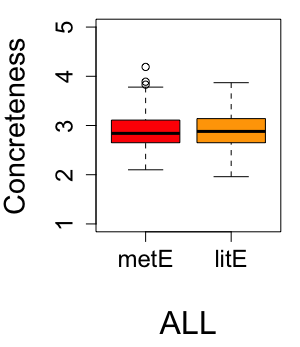} 
\includegraphics[height=50mm]{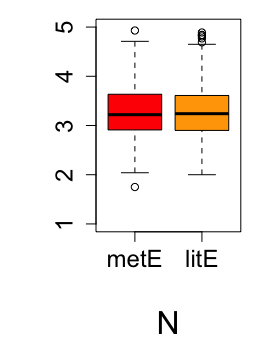} 
\includegraphics[height=50mm]{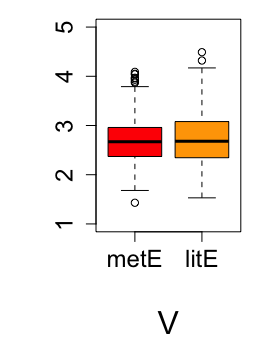} 
\includegraphics[height=50mm]{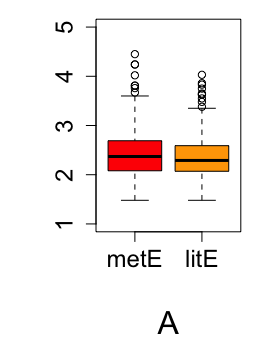} 
\caption{Abstractness of discourse w.r.t metaphorical (metE) vs. literal (litE) \textbf{expressions}, where 1 = more abstract, 5 = more concrete. Ratings are obtained from norms \protect\cite{Brysbaert-etal:2014} based on lexical items provided by annotators: all tokens (ALL), only nouns (N), only verbs (V) and only adjectives (A).}
\label{fig:exps-abstract}
\end{center}
\vspace{-1.9mm}
\end{figure*}

\paragraph{Abstractness} 
We present in Figure~\ref{fig:exps-abstract} the degree of abstractness of the discourses in interaction with the following metaphorical and literal expressions in the respective discourses. We do not observe any indication for metaphorical expressions following more abstract discourses (see red boxes) -- or that literal expressions tend to follow more concrete discourses (see orange boxes). In fact, there does not seem to be a tendency at all that the context's degree of abstractness plays a role in triggering one usage over the other. Our Hypothesis 1b is therefore not confirmed.

\begin{figure}[!h]
\begin{center}
\includegraphics[scale=0.29]{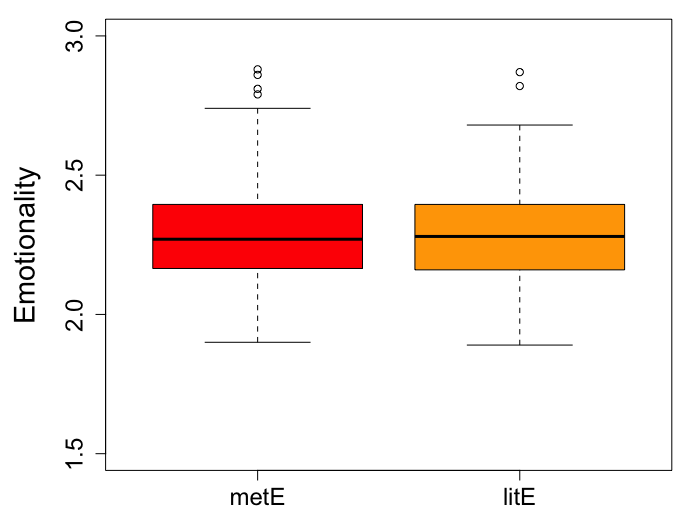} 
\caption{Emotionality of discourses w.r.t metaphorically (metE) vs. literally-perceived (litE) \textbf{expressions}.}
\label{fig:exps-emotion}
\end{center}
\vspace{-2.5mm}
\end{figure}

\paragraph{Emotionality}
Similarly to the abstractness feature, Figure \ref{fig:exps-emotion} suggests that the contexts' degree of emotionality does not seem to play a role in triggering the use of a metaphorical vs. literal expression either, contradicting de facto our Hypothesis 2b. Indeed, even though Figure \ref{fig:discours-emotion} indicated a stronger emotional load of metaphorically-perceived discourses, there is no evidence that a stronger emotional load enforces the use of a metaphorical expression.

\section{Discussion}

Previous research has provided evidence for general ties between abstractness and emotionality with regard to metaphorical language. This work offers the first empirical study on the interaction between abtractness and emotionality with regard to metaphoricity on a discourse level.

Our dataset was annotated differently than in P\&SiW21, as we took a binary rather than a rating scale approach (Section \ref{subsec:experiment}). 
We first looked at whether we obtained the same perceptions in terms of metaphorical vs. literal discourses and observed opposite results (Section \ref{subsec:discourses}). 
These findings suggest the complexity as well as the subjectivity of the task and serve as a reminder to the community that more work is required to explore further ways of how such an annotation study should be performed. Collaborative work needs to be conducted, in order to build reliable gold standard datasets on the perception and properties of metaphorical vs. literal discourses. {\small NLP} applications would greatly benefit from such knowledge: for example, more than simply being able to detect the use of a metaphorical word, a document system summarization should also be able to evaluate how metaphorical the text is in order to preserve the overall perception in the summarized output. 

Our experiments have confirmed interactions of emotionality and abstractness with metaphoricity to a certain extent: overall, metaphorically-perceived discourses are on the one hand more emotional, and on the other hand also more abstract with regard to nouns (Section \ref{subsec:discourses}). These findings thus constitute an important aspect to consider in downstream {\small NLP} applications. 
For instance, most metaphor interpretation systems still approach the task as a paraphrasing task, producing a literal substitution to a metaphorically-used word. \newcite{Mohammad-etal:2016} showed, however, how the emotional content of the text can change when substituting a metaphoric term with a literal paraphrase, and our results have confirmed that this also applies to the discourse level. 
In an application such as machine translation ({\small MT}), we encounter the difficulty to translate figurative language \cite{Volk:1998,Huet-Langlais:2013,Carpuat-Diab:2010,Cholakov-Kordoni:2014} and we observe the partial loss of emotional content \cite{Troiano-etal:2020}.  
In parallel to tackling the challenging task of translating metaphors, one should assure that the abstractness as well as the emotional content loaded in the source text is properly transferred in the target text. Informing a {\small MT} system with an abstractness and/or an emotional classifier could, for example, preserve the abstractness/emotionality content and would therefore favor a metaphorically-perceived discourse in the target language as well.

The second part of the paper focused on the extent to which the metaphoricity, abstractness and emotionality features of the discourses might play a role in triggering the use of a synonymous metaphorical vs. literal expression (Section~\ref{subsec:expressions}). 
Our findings did not support the context-salient hypothesis: a more metaphorical, abstract or emotional discourse does not entail the use of a metaphorical expression. From an {\small NLP} perspective, this suggests that a metaphor generation system does not have to rely on these aspects to preserve consistency and coherence throughout its output, at least not with regard to the ways we implemented our analyses. 

\section{Conclusion}
Previous research has provided evidence for an interaction between abstractness and emotionality with regard to metaphorically-used words and phrases. 
In this paper, we explored these ties on the discourse-level. We first looked at whether abstractness and emotionality of the words in the discourses represent lexical conditions which potentially influence the perception of a given discourse as more or less metaphorical. We then explored whether these features, along with metaphoricity, influence the choice between a metaphorical vs. a synonymous literal expression. 

Our findings align with those of P\&SiW21 and counter the theoretical context-salient hypothesis that metaphorical vs. literal language usage is expected to be preceded by metaphorical vs. literal and/or abstract vs. concrete and/or emotional vs. non-emotional preceding contexts, respectively. 
Our collection however confirms that metaphorically-perceived discourses are overall more emotional, and more abstract with regard to nouns than literally-perceived discourses.

Future work should explore further lexical conditions, such as word classes, complexity and formality, in order to deepen the understanding of metaphorical language usage in larger contexts.

\section*{Acknowledgements}

This research was supported by the DFG Research Grant SCHU 2580/4-1 \textit{Multimodal Dimensions and Computational Applications of Abstractness}.

\section{Bibliographical References}\label{reference}

\bibliographystyle{lrec2022-bib}
\bibliography{lrec2022.bib}

\clearpage
\appendix

\section{Crowdsourcing}
\label{app:crowd}

\hrule
\vspace{.2cm}
\textbf{Metaphoricity/literalness of a text}\\
In this task, we are collecting your judgement on the metaphoricity or literalness of the text.\\

\textbf{Before you start:}\\
Native speakers of English only.\\
Please complete all 20 HITs.\\
Your answer will be approved only if you answer both questions fully.\\

\textbf{Your task:}

Read the whole text attentively. (Note: the data was collected as is (ukWac) from blog posts, comments, etc. It might contain spelling and/or grammatical mistakes: please ignore them.)\\
\underline{Question 1}: Is the text more literal or more metaphorical?\\
\underline{Question 2}: Which words are triggering this decision? Write down 5 unique words which triggered your choice.\\

\textbf{Text:}

\textit{For her, writing is an effective tool to express your viewpoints... To write is already to choose, thus, writing should be done along with a critical mind and a caring soul. She hopes to become more professional, skilled and mature in her craft. Aside from writing Kay spends her time reading. Reading lets her travel to far-off imagined places and situations. She also learns a lot from [...]}\\

\textbf{Task 1: Is the overall text more literal or more metaphorical?}
\vspace{+2mm}\\
$\ocircle$ Literal $\ocircle$ Metaphorical \\

\textbf{Task 2: Which words triggered your choice?}\\
Pick 5 unique words (only 1 word per line. Fill out ALL lines for your answer to be approved.)
\vspace{+2mm}\\
word 1: \_\_\_\_\_\_\_\_\_\_\_\_\_\_\_\_\_\_\_\_\_\_\_\_ \\
word 2: \_\_\_\_\_\_\_\_\_\_\_\_\_\_\_\_\_\_\_\_\_\_\_\_ \\
word 3: \_\_\_\_\_\_\_\_\_\_\_\_\_\_\_\_\_\_\_\_\_\_\_\_ \\
word 4: \_\_\_\_\_\_\_\_\_\_\_\_\_\_\_\_\_\_\_\_\_\_\_\_ \\
word 5: \_\_\_\_\_\_\_\_\_\_\_\_\_\_\_\_\_\_\_\_\_\_\_\_ \\

\vspace{+1mm}
\section{Examples}
\label{app:examples}

\vspace{+2mm}
\subsection{Lexical Items}
\label{appsubsec:words}
We provide a list of the 50 words which were most frequently chosen across discourses by the annotators regarding metaphorical vs. literal discourse perception. This gives an idea of the words that are clearly triggering a metaphorical vs. literal perception of a text.

\begin{table}[!h]
\centering
\begin{tabular}{lr|lr}
    \hline
    \multicolumn{2}{c}{\textbf{Metaphorical}} & \multicolumn{2}{c}{\textbf{Literal}}\\ \hline\hline
    god &   44  &   government  &   149\\
    like    &   30  &   information &   106\\
    life    &   20  &   research    &   89\\
    sin &   19  &   evidence    &   83\\
    think   &   14  &   political   &   77\\
    power   &   14  &   work    &   71\\
    stories &   14  &   results &   68\\
    spiritual   &   13  &   war &   64\\
    freedom &   13  &   study   &   61\\
    human   &   13  &   policy  &   58\\
    mind    &   13  &   data    &   56\\
    spirit  &   12  &   students    &   56\\
    feel    &   11  &   published   &   56\\
    time    &   11  &   people  &   54\\
    peace   &   11  &   development &   53\\
    fight   &   11  &   agreement   &   52\\
    bible   &   10  &   business    &   49\\
    thinking    &   10  &   system  &   45\\
    experience  &   10  &   disease &   45\\
    longing &   10  &   community   &   45\\
    sense   &   10  &   project &   45\\
    hope    &   9   &   report  &   45\\
    heart(s)  &   9   &   history &   43\\
    myth    &   9   &   university  &   43\\
    jesus   &   9   &   decision    &   42\\
    fate    &   9   &   council &   41\\
    ghost   &   9   &   education   &   41\\
    drunk   &   9   &   experience  &   41\\
    dragon  &   8   &   response    &   39\\
    world   &   8   &   support &   38\\
    emotional   &   8   &   analysis    &   38\\
    cultural    &   8   &   example &   37\\
    dare    &   8   &   said    &   37\\
    imagination &   8   &   important   &   36\\
    memory  &   8   &   public  &   36\\
    angels  &   8   &   survey  &   36\\
    satan   &   8   &   working &   35\\
    artistic    &   8   &   review  &   35\\
    lord    &   8   &   have    &   34\\
    beyond  &   8   &   money   &   33\\
    value   &   8   &   economic    &   33\\
    drifting    &   8   &   can &   32\\
    oblivion    &   8   &   need    &   32\\
    spell   &   8   &   assessment  &   32\\
    perception  &   8   &   school  &   31\\
    veneer  &   8   &   legal   &   31\\
    chips   &   8   &   industry    &   31\\
    energy  &   7   &   debate  &   31\\
    empowerment &   7   &   skills  &   30\\
    scriptures  &   7   &   official    &   29\\
    \hline
\end{tabular}
\caption{Most picked words by the annotators which triggered them to perceive a discourse as more metaphorical (left) or more literal (right), and the number of times those respective words were provided.}
\label{tab:words}
\end{table}

\onecolumn
\subsection{Agreement on Metaphoricity across Annotation Studies (Anno1\&2)}
\label{appsubsec:intersec}
We provide below two examples each for discourses that were perceived as metaphorical and literal in both P\&SiW21 (Anno1) and the present's (Anno2) annotation studies. Additionally, we show in italics the lexical items that triggered at least two annotators in the present study to make their decision regarding the metaphoricity of the discourses. Note that several of these lexical items are amongst the most picked words presented in Table \ref{tab:words}.

\vspace{+3mm}
\textbf{Metaphorical Discourses:}
\vspace{+2mm}\\
(1) The only way left over to you to make your God \textit{impartial} is that you must accept that your \textit{God} appeared in all the countries at a time in various forms and \textit{preached} your path in various languages. The same \textit{form} did not \textit{appear} everywhere and the same language does not exist everywhere. The \textit{syllabus} and \textit{explanation} are one and the same, though the \textit{media} and \textit{teachers} are different. Can you give any alternative \textit{reasonable} answer to my question other than this? Certainly not! Any person of any religion to any other \textit{religion} can [...]
\vspace{+2mm}\\
(2) What is more, they have \textit{retained} their \textit{emotional}, \textit{cultural} and \textit{spiritual} links with the country of their origin. This strikes a \textit{reciprocal} \textit{chord} in the \textit{hearts} of people of India. Also, any \textit{longed-for} return to the homeland now tends to be downplayed in favor of \textit{ideological} identification or transnational practice that can link the scattered community with the homeland. Today, \textit{self-defined} \textit{diasporas} tend to find esteem -- and a kind of strength-in-numbers -- through using the term. This shift in the adoption and meaning of "diaspora" has undoubtedly caused some confusion and [...]

\vspace{+3mm}
\textbf{Literal Discourses:}
\vspace{+2mm}\\
(1) We will enable local \textit{authorities} to \textit{purchase} land \textit{compulsorily}, where the land is the subject of a continuing \textit{breach} of a Stop Notice. And we will allow councils to refuse applications for \textit{retrospective} planning permission where it is clear that the \textit{applicant} knew they were breaking the law. As always our policy will be to \textit{enhance} and not diminish the power that rural communities have to run themselves through their parish, district and county councils. Rural \textit{policy} is complex which is another reason why, under a Conservative \textit{Government}, \textit{centralisation} will be out and local \textit{democracy} in. But until then the [...]
\vspace{+2mm}\\
(2) 5 years ago in Tottenham, we began to \textit{focus} more of our energies on the attainment levels of our young people, and GCSE \textit{results} have improved for the fifth year in succession, and the proportion of \textit{students} achieving 5 A* to C grades at GCSE has hit 50\% for the first time in Haringey's \textit{history}, up from 31\% five years ago. It needs to be higher. \textit{Improvement} by African Caribbeans at GCSE rose by more the national \textit{average} last year, but it needs to improve faster every year until we have eliminated the achievement gap altogether. Many of us in this room today recognise the difficult \textit{environment} that these young \textit{people} describe all too well. Why? Because we grew up in it. We know that however many of us have gone on to succeed in what we wanted to achieve, to [...]

\end{document}